\documentclass[sigconf,screen]{acmart}
\usepackage{multirow}

\copyrightyear{2023}
\acmYear{2023}
\setcopyright{acmlicensed}\acmConference[BuildSys '23]{The 10th ACM International Conference on Systems for Energy-Efficient Buildings, Cities, and Transportation}{November 15--16, 2023}{Istanbul, Turkey}
\acmBooktitle{The 10th ACM International Conference on Systems for Energy-Efficient Buildings, Cities, and Transportation (BuildSys '23), November 15--16, 2023, Istanbul, Turkey}
\acmPrice{15.00}
\acmDOI{10.1145/3600100.3623726}
\acmISBN{979-8-4007-0230-3/23/11}

\begin{document}

\title[OoD Electricity Forecasting during COVID-19]{
Navigating Out-of-Distribution Electricity Load Forecasting during COVID-19: Benchmarking energy load forecasting models without and with continual learning
}

\author{Arian Prabowo}
\email{arian.prabowo@unsw.edu.au}
\orcid{0000-0002-0459-354X}
\affiliation{%
  \institution{University of New South Wales}
  \city{Sydney}
  \country{Australia}
}

\author{Kaixuan Chen}
\email{kaixuan.chen@unsw.edu.au}
\orcid{0000-0003-3904-0395}
\affiliation{%
  \institution{University of New South Wales}
  \city{Sydney}
  \country{Australia}
}

\author{Hao Xue}
\email{hao.xue1@unsw.edu.au}
\orcid{0000-0003-1700-9215}
\affiliation{%
  \institution{University of New South Wales}
  \city{Sydney}
  \country{Australia}
}

\author{Subbu Sethuvenkatraman}
\email{subbu.sethuvenkatraman@csiro.au}
\orcid{0000-0001-7197-2307}
\affiliation{%
  \institution{CSIRO}
  \city{Newcastle}
  \country{Australia}
}

\author{Flora D. Salim}
\email{flora.salim@unsw.edu.au}
\orcid{0000-0002-1237-1664}
\affiliation{%
  \institution{University of New South Wales}
  \city{Sydney}
  \country{Australia}
}

\renewcommand{\shortauthors}{Prabowo et al.}

\begin{abstract}
In traditional deep learning algorithms, one of the key assumptions is that the data distribution remains constant during both training and deployment.
However, this assumption becomes problematic when faced with Out-of-Distribution periods, such as the COVID-19 lockdowns, where the data distribution significantly deviates from what the model has seen during training.
This paper employs a two-fold strategy: utilizing continual learning techniques to update models with new data and harnessing human mobility data collected from privacy-preserving pedestrian counters located outside buildings.
In contrast to online learning, which suffers from 'catastrophic forgetting' as newly acquired knowledge often erases prior information, continual learning offers a holistic approach by preserving past insights while integrating new data.
This research applies FSNet, a powerful continual learning algorithm, to real-world data from 13 building complexes in Melbourne, Australia, a city which had the second longest total lockdown duration globally during the pandemic.
Results underscore the crucial role of continual learning in accurate energy forecasting, particularly during Out-of-Distribution periods.
Secondary data such as mobility and temperature provided ancillary support to the primary forecasting model.
More importantly, while traditional methods struggled to adapt during lockdowns, models featuring at least online learning demonstrated resilience, with lockdown periods posing fewer challenges once armed with adaptive learning techniques.
This study contributes valuable methodologies and insights to the ongoing effort to improve energy load forecasting during future Out-of-Distribution periods.
\end{abstract}



\begin{teaserfigure}
\includegraphics[width=\textwidth]{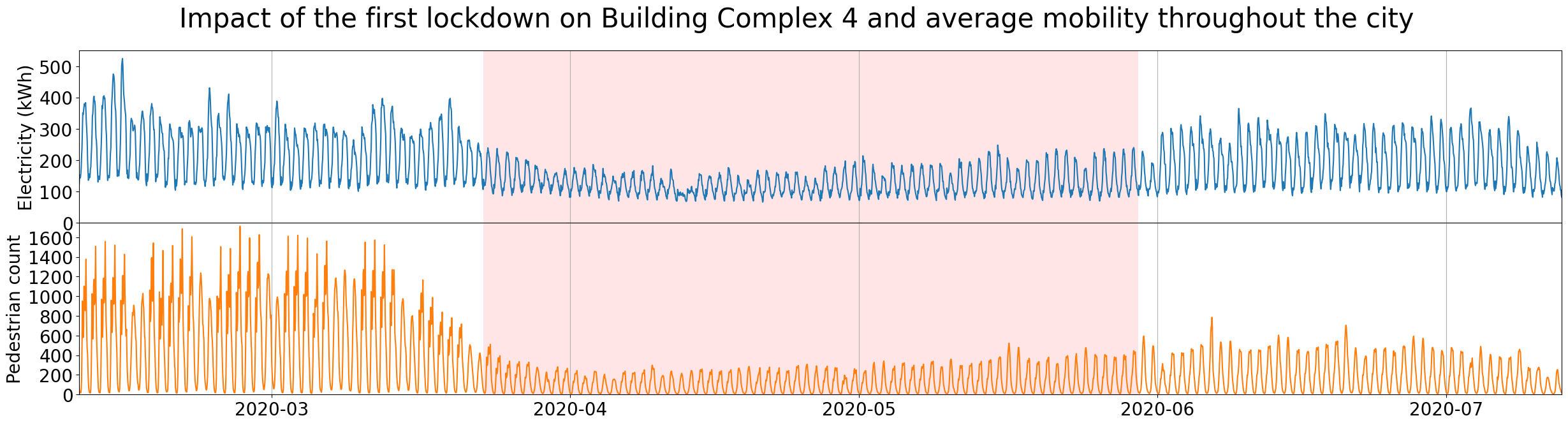}
\caption{
This figure illustrates the impact of the first COVID-19 lockdown (indicated by the red region) on the electricity load of a building complex located in Melbourne's Central Business District (top graph).
Concurrently, it presents the effect on the average human mobility throughout the city, as measured by pedestrian count (bottom graph).
The diagram clearly shows significant shifts in distributions during the lockdown, which do not fully revert to their pre-lockdown state even after restrictions are lifted.
}
\label{fig:vizabs}
\end{teaserfigure}


\maketitle

\section{Introduction}

Buildings play a significant role in energy consumption, making accurate electricity load forecasting essential for effective energy management and maintaining a balance in power grids. 
However, accurate forecasting is a challenging task since the building energy usage is primarily influenced by behaviours or activities of the occupants that are highly dynamic and heterogeneous~\cite{CHEN2021101212}.
This can be exacerbated during Out-of-Distribution periods, such as the COVID-19 pandemic and lockdowns, presenting unique challenges due to changes in occupancy patterns and energy usage behavior.
During these anomalous periods, we show that traditional machine learning and deep learning methods struggle to adapt, highlighting the need for different approaches that can effectively capture and account for shifting dynamics.

\subsection{Online and Continual Learning}
In conventional timeseries forecasting using machine learning, models are trained on a fixed training dataset and stop learning once deployed. Consequently, these models cannot adapt to changes in underlying data patterns during deployment.
Online Learning (OL) has emerged as a promising solution to this challenge, as it enables building managers to adapt to changes in occupancy patterns and adjust energy usage accordingly.
With OL, models can be updated incrementally with each new data point, allowing them to learn and adapt in real-time~\cite{hoi2021OLsurvey}.
Furthermore, a subset of OL method called continual learning offers an even more powerful solution by addressing the issue of catastrophic forgetting~\cite{lin1992ER,chaudhry2019ER}.
These methods allow models to retain previously learned information while accommodating new data, preventing the loss of valuable insights and improving generalization in out-of-distribution scenarios.
With continual learning technique, energy forecasting models can achieve robustness, adaptability, and accuracy, making them well-suited for handling the challenges posed by spatiotemporal data with evolving distributions.

\subsection{Human Mobility as a Proxy to Occupancy Behaviour}
Moreover, accurately forecasting buildings electricity load is challenging due to the dynamic and diverse nature of occupant behaviors, which are often not directly observable.
Occupancy data is crucial~\cite{HOSSEINI2022103634,anand2021occupancy} but difficult to obtain due to privacy concerns, the absence of occupancy sensors in legacy buildings, high sensor installation costs, and issues related to sensor accuracy.
Researchers have explored various proxy information sources for occupancy sensing, such as
Google search trends~\cite{FU2022118343},
positional records from Twitter~\cite{MOHAMMADI2017810},
Call Details Recore \cite{Bogomolov2016-fq},
Wi-Fi-derived occupancy~\cite{Nweye},
Bluetooth low energy on mobile devices~\cite{TEKLER2020106681},
domestic water consumption~\cite{Fu2021-ry},
and indoor carbon dioxide measurements~\cite{Yixuan}.
However, there has been limited focus on leveraging human mobility-related spatial-temporal data for energy use modeling~\cite{Miller}.
While occupancy sensors provide complete observation of people in buildings, pedestrian sensors only capture partial pedestrian flows, making the utilization of traffic data for energy modeling more challenging.
In this study, we investigate the extent to which we can leverage pedestrian sensors data to improve electricity load forecasting.

\subsection{Contributions}
In this study, we evaluate the effectiveness of mobility data and continual learning for forecasting buildings electricity load during out-of-distribution periods.
We utilized real-world data from Melbourne, Australia, a city that experienced one of the strictest lockdowns globally~\cite{boaz_2021}, making it an ideal case study.
We conducted experiments using data from 13 building complexes to empirically assess the performance of these methods.
Our contributions are as follows:
\begin{enumerate}
    \item Our dataset provides a high spatial granularity, covering 13 buildings in Melbourne, Australia, which experienced one of the longest and strictest lockdowns globally.
    \item We demonstrate the superiority of continual learning over traditional OL and non-OL methods, particularly during out-of-distribution periods. Classical machine learning  and deep learning methods experience poor performance during out of distribution scenarios.
    \item Incorporating mobility data as a feature holds potential for enhancing forecasting accuracy, but its applicability should be carefully evaluated to prevent overfitting.
    \item Our study reveals that forecasting during lockdown periods was surprisingly easier than pre-lockdown periods, but this observation is valid only when employing online or continual learning approaches.
\end{enumerate}

\section{Related Works}

\subsection{Energy Prediction in Urban Environments}
Electricity demand profiling and forecasting has been a task of importance for many decades. Nevertheless, there exist a limited number of work in literature that investigate how human mobility patterns are directly related to the urban scale energy consumption, both during normal periods as well as adverse/extreme events. Energy modelling in literature is done at different granularities, occupant-level (personal energy footprinting), building-level and city-level. Models used for energy consumption prediction in urban environments are known as Urban Building Energy Models (UBEM). While top-down UBEMs are used for predicting aggregated energy consumption in urban areas using macro-economic variables and other aggregated statistical data, bottom-up UBEMs are more suited for building-level modelling of energy by clustering buildings into groups of similar characteristics~\cite{ALI2021111073}. Some examples in this respect are SUNtool, CitySim, UMI, CityBES, TEASER and HUES. Software modelling (simulation-based) is also a heavily used approach for building-wise energy prediction (Eg: EnergyPlus~\cite{CRAWLEY2001319}). Due to fine-grain end-user level modelling, bottom-up UBEMs can incorporate inputs of occupant schedules. There also exist occupant-wise personal energy footprinting systems. However, for such occupant-wise energy footprinting, it requires infrastructure related to monitoring systems and sensors for indoor occupant behaviours, which are not always available. Also, due to privacy issues, to perform modelling at end-user level granularity, it can be hard to get access to publicly available data at finer temporal resolutions (both occupancy and energy)~\cite{Wei2019-hw}. Building-wise energy models also have the same problems. Simulation-based models have complexity issues when scaling to the city level, because they have to build one model per each building. Moreover, simulation-based models contain assumptions about the data which make their outputs less accurate~\cite{Ali2020-me}. Consequently, it remains mostly an open research area how to conduct energy forecasting with data distribution shifts.

\subsection{Mobility Data as Auxiliary Information in Forecasting}
The study of human mobility patterns involves analysing the behaviours and movements of occupants in a particular area in a spatio-temporal context~\cite{Salim2020-fc}. The amount of information that mobility data encompasses can be huge. The behaviour patterns of humans drive the decision making in many use-cases. Mobility data in particular, can act as a proxy for the dynamic (time varying) human occupancy at various spatial densities (building-wise, city-wise etc.). Thus such data are leveraged extensively for many tasks in urban environments including predicting water demand~\cite{Smolak2020-xc}, urban flow forecasting~\cite{Xue2021-br, xue2021mobtcast}, predicting patterns in hospital patient rooms~\cite{DEDESKO2015136}, electricity use~\cite{hansika2023human} etc.\ that depend on human activities.

Especially, during the COVID19 pandemic, mobility data has been quite useful for disease propagation modelling. For example, in the work by \cite{Wang2020-hi}, those authors have developed a Graph Neural Network (GNN) based deep learning architecture to forecast the daily new COVID19 cases state-wise in United States. The GNN is developed such that each node represents one region and each edge represents the interaction between the two regions in terms of mobility flow. The daily new case counts, death counts and intra-region mobility flow is used as the features of each node whereas the inter-region mobility flow and flow of active cases is used as the edge features. Comparisons against other classical models which do not use mobility data has demonstrated the competitiveness of the developed model.

Nevertheless, as \cite{Salim2020-fc} state, the existing studies involving human mobility data lack diversity in the datasets in terms of their social demographics, building types, locations etc. Due to the heterogeneity, sparsity and difficulty in obtaining diverse mobility data, it remains a significant research challenge to incorporate them in modelling techniques~\cite{ALI2021111073}. Yet, the lack of extracting valuable information from such real-world data sources remains untapped, with a huge potential of building smarter automated decision making systems for urban planning~\cite{Salim2020-fc}.

\subsection{Deep Online and Continual Learning}
Deep learning methods have been used for many mobility data and tasks, such as flight delay prediction \cite{shao2022predicting}, map inference \cite{prabowo2019COLTRANE} and time-series forecasting \cite{prabowo2023GSWaN,prabowo2023MPNN4TrafficForecasting,prabowo2023unseen_roadsSCPT,liu2023self}.
Moreover, methods such as Long Short-Term Memory (LSTM)~\cite{pelka2020lstmLoadForecasting}, Neural basis expansion analysis for interpretable time series forecasting (N-BEATS)~\cite{ORESHKIN2021116918NBEATSenergyForecasting} and transformers \cite{xue2023promptcast} have been widely used for electricity load forecasting.
However, a common challenge for deep learning methods is the performance degradation during out-of-distribution periods.
Online learning methods have been proposed to address this, but they can suffer from catastrophic forgetting~\cite{Salim2020-fc}.
To mitigate this, continual learning methods such as Fast and Slow Network (FSNet)~\cite{pham2022learning} have been developed.
These methods aim to retain past knowledge while adapting to new data, improving generalization in OOD scenarios.

\section{Datasets}

In this section, we will discuss the sources and composition of the datasets used in our research.
This includes the electricity load dataset,
the division of our timeline into five distinctive periods aligning with the lockdowns due to COVID-19,
the human mobility data captured by automated pedestrian counting systems,
and the temperature data.

\begin{table*}[htb]
\centering
\caption{
The summary statistics of the energy dataset from 13 anonymized building complexes (BCs).
}
\label{tab:bc}
\begin{tabular}{@{}c|ccc|rr|rrrrr@{}}
\toprule
\textbf{BC} & \multicolumn{1}{c}{\textbf{start}} & \multicolumn{1}{c}{\textbf{end}} & \multicolumn{1}{c|}{\textbf{\begin{tabular}[c]{@{}c@{}}duration\\ (years)\end{tabular}}} & \multicolumn{1}{c}{\textbf{\begin{tabular}[c]{@{}c@{}}mean\\ (kWh)\end{tabular}}} & \multicolumn{1}{c|}{\textbf{\begin{tabular}[c]{@{}c@{}}std.\\ (kWh)\end{tabular}}} & \multicolumn{1}{c}{\textbf{\begin{tabular}[c]{@{}c@{}}min\\ (kWh)\end{tabular}}} & \multicolumn{1}{c}{\textbf{\begin{tabular}[c]{@{}c@{}}0.25\\ (kWh)\end{tabular}}} & \multicolumn{1}{c}{\textbf{\begin{tabular}[c]{@{}c@{}}median\\ (kWh)\end{tabular}}} & \multicolumn{1}{c}{\textbf{\begin{tabular}[c]{@{}c@{}}0.75\\ (kWh)\end{tabular}}} & \multicolumn{1}{c}{\textbf{\begin{tabular}[c]{@{}c@{}}max\\ (kWh)\end{tabular}}} \\ \midrule
\textbf{1}  & 2018-08-01  & 2020-12-31 & 2.42 & 276.73 & 99.68  & 19.38 & 213.31 & 280.78 & 347.45 & 686.01 \\
\textbf{2}  & 2019-02-13  & 2020-12-31 & 1.88 & 12.33  & 6.68   & 0.00  & 8.50   & 10.31  & 13.65  & 50.16  \\
\textbf{3}  & 2019-01-01  & 2020-12-31 & 2.00 & 207.27 & 111.59 & 3.25  & 112.72 & 169.29 & 297.33 & 611.67 \\
\textbf{4}  & 2019-01-01  & 2020-12-31 & 2.00 & 223.27 & 108.37 & 0.00  & 134.52 & 199.20 & 307.09 & 611.18 \\
\textbf{5}  & 2019-01-01  & 2020-12-31 & 2.00 & 12.94  & 9.65   & 3.68  & 5.77   & 8.85   & 17.65  & 55.26  \\
\textbf{6}  & 2018-01-01  & 2020-12-31 & 3.00 & 278.17 & 88.67  & 1.32  & 203.38 & 272.68 & 342.10 & 709.41 \\
\textbf{7}  & 2018-01-01  & 2020-12-31 & 3.00 & 166.42 & 66.60  & 5.38  & 112.62 & 144.34 & 206.63 & 371.64 \\
\textbf{8}  & 2018-01-01  & 2020-12-31 & 3.00 & 160.09 & 96.24  & 11.25 & 92.68  & 120.79 & 217.45 & 550.47 \\
\textbf{9} & 2019-01-01  & 2020-12-31 & 2.00 & 72.83  & 33.64  & 1.80  & 45.18  & 66.71  & 96.57  & 218.07 \\
\textbf{10}  & 2019-01-01  & 2020-12-31 & 2.00 & 34.48  & 29.20  & 5.51  & 14.80  & 21.76  & 41.00  & 123.90 \\
\textbf{11} & 2019-07-01  & 2020-12-31 & 1.50 & 26.64  & 13.21  & 0.00  & 17.45  & 21.75  & 31.06  & 83.01  \\
\textbf{12} & 2019-01-01  & 2020-12-31 & 2.00 & 0.90   & 1.04   & 0.20  & 0.28   & 0.62   & 0.96   & 15.08  \\
\textbf{13} & 2018-01-01  & 2020-10-31 & 2.83 & 293.53 & 96.17  & 0.23  & 231.14 & 286.68 & 351.01 & 675.15 \\
\bottomrule
\end{tabular}
\end{table*}

\subsection{Electricity Load}

This research relies on electricity data obtained from electricity suppliers, with privacy being a paramount consideration.
To protect the privacy of building owners, operators, and users, nearby buildings are aggregated into anonymized Building Complexes (BCs).
These BCs consist of mixed-use properties, encompassing residential, office, and retail spaces.
The dataset under analysis is derived from thirteen such BCs located around Melbourne's Central Business District, Australia.
four of these BC datasets have been previously published \cite{prabowo2023continually}.

A statistical summary of the data is provided in Table \ref{tab:bc} and its visual representation can be found in Figure 1.
The dataset maintains an hourly granularity, presenting a detailed view of energy consumption.
It is noteworthy that this dataset displays considerable variation in scale, with the mean value of BC 13 being about 300 times that of BC 12.

\subsection{Lockdown Time Periods} \label{sec:periods}

The Melbourne lockdown timelines are complex due to the varying levels of restrictions imposed at different stages. Consequently, it's not a simple binary state of lockdown versus no lockdown. Despite these complexities, for the purpose of our study, we've categorized the timeline into five distinct periods:
\begin{itemize}
    \item[\textbf{PLD}] Pre-Lockdown: The period before the first lockdown.
    \item First lockdown starts on 23rd March 2020~\cite{Parliamentary_Library_Campbell_Vines_2021}.
    \item[\textbf{LD1}] First Lockdown.
    \item First lockdown ends on 30th May 2020.
    \item[\textbf{IL1}] Inter-Lockdown One: The period between the end of the first lockdown and the start of the second lockdown.
    \item Second lockdown starts on 2nd August 2020~\cite{Andrew_2020}.
    \item[\textbf{LD2}] Second Lockdown.
    \item Second lockdown ends on 27th October 2020~\cite{Andrew_2020b}.
    \item[\textbf{IL2}] Inter-Lockdown Two: The period from the end of the second lockdown until the end of 2021, which concludes our datasets.
\end{itemize}

\subsection{Human Mobility Data}

Mobility data was obtained from an automated pedestrian counting system deployed by the City of Melbourne, which tracked pedestrian movements throughout the city without collecting any personal information, as no video was recorded.
From the 79 installed sensors, we chose 42 that provided complete coverage over our desired timeframe, taking into account the varying installation times, as well as instances of relocation and removal.
A detailed description of this data is not provided herein as they are publicly accessible \cite{CoM_peds}.

\subsection{Temperature}

We sourced temperature records from the National Renewable Energy Laboratory (NREL)~\cite{nrel_data} Asia Pacific Himawari Solar Data. Given the close proximity of the BCs, the same temperature data was utilized for all.

\section{Experiments}

\subsection{Problem Definition}

We define the forecasting problem as follows.
Given the recent past observation
$\mathbf{X} \in \mathbb{R}^{W \times |F|}$
where
$W \in \mathbb{Z}^{+}$ is the length of the observation window
and
$F$ is the feature set (descibed later),
the task to forecast the electricity load in the near future
$\mathbf{Y} \in \mathbb{R}^{H}$
where
$H$ is the forecasting horizon.
Thus,
$f(\mathbf{X}) = \mathbf{Y}$.

\subsection{Online Learning Setup}

Our experimental setup is designed to evaluate the model's ability to generalize during Out-of-Distribution periods.
We adopt a time-based splitting strategy to divide the dataset.
The model's initial training is performed on the first three months of the data, referred to as the 'warm-up' period.

After this initial setup, we incorporate two separate strategies for the rest of the data:
\textbf{Classical}: In this approach, no further training is done post the warm-up period. The model, once trained, is used as is to forecast for the remaining period.
\textbf{Online Learning (OL)}: In this method, the model undergoes a gradient descent step for every new incoming data point i.e. $\mathtt{batch\_size}=1$. It implies that the model learns and updates sequentially from each new data point.

For hyperparameter tuning, the validation set consists of one month of data following the three months used for training / warm-up.

\subsection{Feature Sets}

We explore the role of human mobility in comparison with temperature in forecasting by testing four different feature sets ($F$).
This is done to understand how these variables contribute to the model's performance. 

\begin{itemize}
    \item[E] $|F_1| = 3 = 1\ E + 1\ HoD + 1\ DoW$ where $E$ is Energy Use, $HoD$ is the hour-of-day, and $DoW$ is the day-of-week.
    \item[EM] $|F_2|=47=|F_1|+42\ M + 1\ \bar{M}_{micro} + 1\ \bar{M}_{macro}$ where each $M$ is pedestrian counter, $\bar{M}_{micro}$ is the micro-average of the pedestrian counter, and $\bar{M}_{micro}$ is the macro-average of the pedestrian counter. Micro-average is the usual average, while in macro-average, each pedestrian counter timeseries is [0,1] normalized first before averaging.
    \item[ET] $|F_3|=4=|F_1|+1\ T$ where $T$ is temperature.
    \item[ETM] $|F_4|=48=|F_2|+1\ T$ uses all of the features.
\end{itemize}

\subsection{Selected Methods}

\subsubsection{Naive}
In this study, we employ 11 baseline methods, beginning with three naive approaches.
In the \texttt{CopyLastHour} method, forecasts are simply generated by duplicating the last hour's value, presuming smoothness in the time series.
This is a common naive baseline in forecasting tasks \cite{prabowo2023MPNN4TrafficForecasting,jiang2021dl}.
Similar in approach, \texttt{CopyLastDay} and \texttt{CopyLastWeek} are methods that replicate values from the previous day and week respectively, thus assuming periodic behaviors within the time series.

\subsubsection{Statistical}
Transitioning from the naive methods, we utilize Exponential Smoothing (\texttt{ES}).
This method is a typical statistical non-learning method that, despite its simplicity, does involve a hyperparameter that we validate using the first month of data.
Considering the rapid computational ability of \texttt{ES}, a thorough grid search is performed through all datasets for this hyperparameter, which is the smoothing factor ($\alpha$), ranging from $0<\alpha<1$ with a step size of $0.0001$.
It is important to note that the first four methods are non-learning and univariate, and do not conform to the setup described earlier.
We discovered that the best $\alpha$ is 1, thus ES is identical with \texttt{CopyLastHour}.

\subsubsection{Machine Learning}

Random Forest (\texttt{RF}) regression is a common machine learning algorithm \cite{breiman2001randomForest,geurts2006extremelyRF}.
We use the implementation by \url{www.scikit-learn.org} version 1.2.2.
The hyperparameter search space is uniformly distributed as follows:
\begin{itemize}
    \item \texttt{n\_estimators}, range=[1,200]
    \item \texttt{max\_depth}, range=[1,50]
    \item \texttt{min\_samples\_split}, range=[0,1]
    \item \texttt{min\_samples\_leaf}, range=[0,1]
    \item \texttt{min\_weight\_fraction\_leaf}, range=[0,0.5]
    \item \texttt{max\_features}, range=[0,1]
    \item \texttt{max\_leaf\_nodes}, range=[2,50]
    \item \texttt{oob\_score}, range=[0,1]
    \item \texttt{max\_samples}, range=[0,1]
\end{itemize}

\begin{figure*}[htb]
\centering
\includegraphics[width=\linewidth]{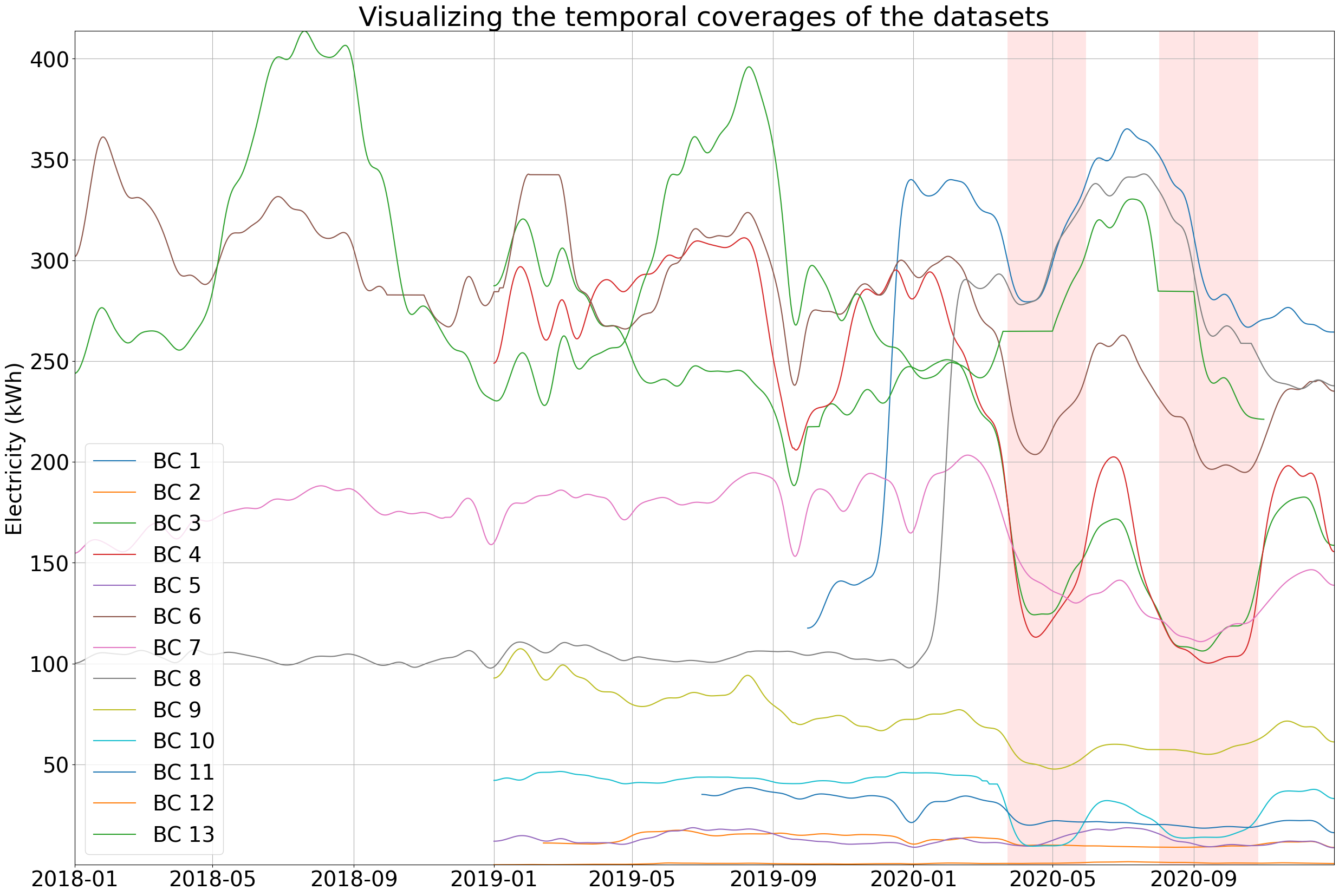}
\caption{
The visualization showcases electricity datasets from thirteen Building Complexes (BCs), emphasizing the inherent variations in electricity load across these BCs.
It also highlights the significant changes in load patterns triggered by lockdowns.
To facilitate the visualization, the plots smoothed using a Gaussian filter with a sigma of one week.
}
\label{fig:data_electricity}
\end{figure*}

eXtreme Gradient Boosting (\texttt{XGBoost}) is another popular machine learning algorithm \cite{chen2016xgboost}.
It can handle many different task, and is also popular for timeseries forecasting \cite{Herzen2022Darts}.
We use the official implementation at \url{xgboost.readthedocs.io/en/stable/python/python_intro.html} version 1.7.5.
The hyperparameter search space is as follows:
\begin{itemize}
    \item \texttt{max\_depth}, range = [1, 10], uniform distribution
    \item \texttt{min\_child\_weight}, range = [1e-4, 1e4], logarithmic distribution
    \item \texttt{gamma}, range = [1e-4, 1e4], logarithmic distribution
    \item \texttt{subsample}, range = [1e-8, 1.0], logarithmic distribution
    \item \texttt{colsample\_bytree}, range = [1e-8, 1.0], logarithmic distribution
    \item \texttt{eta}.3, range = [1e-8, 1.0], logarithmic distribution
    \item \texttt{alpha}, range = [1e-8, 1.0, logarithmic distribution]
\end{itemize}

\texttt{VAR} is a common baseline for multi-variate timeseries \cite{prabowo2023GSWaN}.
We found differencing to be not helpful.

\subsubsection{Deep Learning} \label{sec:baseline_dl}

A typical deep learning baseline for electricity load forecasting is Long Short-Term Memory (\texttt{LSTM})~\cite{pelka2020lstmLoadForecasting,ORESHKIN2021116918NBEATSenergyForecasting}.
We use our own PyTorch implementation.
Following previous work~\cite{ORESHKIN2021116918NBEATSenergyForecasting}, we use a single LSTM layer, 20 epochs without early stopping, gradient clipping at L2 norm of 1.
Because 3 months consist of about 2160 hourly timesteps, we pick a batch size of 64 and ended up with 33 weights updates per epoch.
The hyperparameter search space is as follows:
\begin{itemize}
    \item \texttt{hidden\_size}, range = [50, 1000], uniform distribution.
    \item \texttt{learning\_rate}, range = [1e-10, 1e-2], logarithmic distribution.
\end{itemize}

The generic version of Neural Basis Expansion Analysis for interpretable Time Series (\texttt{NBEATS}) have also been used for electricity load forecasting~\cite{ORESHKIN2021116918NBEATSenergyForecasting}.
the original paper is univariate.
While the original implementation is only for univariate forecasting, we adapt it to handle multivariate inputs by flattening the input data \cite{Herzen2022Darts}.
However, in feature set E, we do not include the time features as the original paper is designed for univariate forecasting.
We use the official PyTorch implementation at \url{www.github.com/ServiceNow/N-BEATS}.
The hyperparameter search space is as follows:
\begin{itemize}
    \item \texttt{stacks}, range = [1, 10], uniform distribution.
    \item \texttt{layers}, range = [2, 4], uniform distribution.
    \item \texttt{layer\_size}, range = [2, 4], uniform distribution.
    \item \texttt{learning\_rate}, range = [1e-10, 1e-2], logarithmic distribution.
\end{itemize}

\subsubsection{Online Learning}

In the \texttt{VAR+OL} method, we apply OL to the VAR model using the Adam optimizer \cite{Kingma2015adam}.
The learning rate is the only hyperparameter we tuned, and we set the range between $1 \times 10^{-6}$ and $5 \times 10^{-2}$.

Similarly, we also apply OL to LSTM and NBEATS to make \texttt{LSTM+OL} and \texttt{NBEATS+OL}, using the same hyperparameters as described in Section 
\ref{sec:baseline_dl}.

\subsubsection{Continual Learning}
Fast and Slow Network (\texttt{FSNet}) \cite{pham2022learning} is a deep continual learning method.
We only search through the hyperparameters recommended in the original paper \cite{pham2022learning} as follows:
\begin{itemize}
    \item $\gamma$               , range = (0, 1), uniform distribution.
    \item $\gamma'$              , range = (0, 1), uniform distribution.
    \item $\tau$                 , range = (0, 1), uniform distribution.
    \item \texttt{learning\_rate}, range = [1e-10, 1e-2], logarithmic distribution.
\end{itemize}

\subsection{Experimental Setups}

In our experimental setup, we apply standard scaling to the data and utilize the Adam optimizer for all deep learning, online learning, and continual learning methods.
We clamp or clip the model's output between the minimum and maximum values in a time period (see Section~\ref{sec:periods}).
The mean absolute error (MAE) serves as the loss function.
While various evaluation metrics are available, our focus primarily lies on showcasing the MAE metrics, as they provide consistent conclusions across different evaluation criteria.
For models with hyperparameters, we employ Optuna~\cite{akiba2019optuna} with a budget of 20 runs to optimize their performance.
Additionally, for stochastic models, we replicate the experiments 10 times with different random seeds and report the mean results.

\subsubsection{Softward and Hardware}
In our experiments, we utilized the following software:
Python 3.9.2,
PyTorch 1.9.0,
Optuna 2.10.0,
NumPy 1.20.0, and
pandas 1.2.4.
The experiments were conducted using the Google Colab PRO platform, with the release date set to June 2, 2023 and
the Gadi high performance computing system \url{www.top500.org/system/179865}.

\begin{table}[htb]
\centering
\caption{
The comprehensive results of all combinations of methods and feature sets for BC 8 are presented, utilizing the MAE as the error metric - where a lower value is indicative of better performance.
The results are arranged in ascending order based on the PLD column, hence the top result in bold with lowest error is the best.
FSNet is a popular model that incorporate continual learning.
}\label{tab:bc8}
\begin{tabular}{@{}cc|rrrrr@{}}
\toprule
\textbf{Methods} & \textbf{Feat.} & \multicolumn{1}{c}{\textbf{PLD}} & \multicolumn{1}{c}{\textbf{LD1}} & \multicolumn{1}{c}{\textbf{IL1}} & \multicolumn{1}{c}{\textbf{LD2}} & \multicolumn{1}{c}{\textbf{IL2}} \\ \midrule \midrule

\textbf{FSNet} & \textbf{EM} & \textbf{5.26} & \textbf{11.79} & \textbf{12.50} & \textbf{10.45} & \textbf{8.10} \\
\textbf{FSNet} & ET & 5.67 & 12.46 & 12.86 & 10.89 & 8.58\\
\textbf{FSNet} & E & 5.76 & 13.59 & 14.44 & 11.81 & 9.09\\
\textbf{NBEATS+OL} & EM & 7.01 & 16.41 & 17.90 & 17.41 & 13.79\\
\textbf{FSNet} & ETM & 7.05 & 14.48 & 14.43 & 11.45 & 8.87\\
\textbf{NBEATS+OL} & E & 7.90 & 15.74 & 18.05 & 16.47 & 13.75\\
\textbf{NBEATS+OL} & ETM & 8.02 & 18.36 & 17.84 & 17.74 & 13.28\\
\textbf{NBEATS+OL} & ET & 8.18 & 19.39 & 18.02 & 16.90 & 12.78\\
\textbf{LSTM+OL} & ETM & 10.21 & 22.17 & 20.99 & 18.31 & 13.22\\
\textbf{LSTM+OL} & EM & 10.75 & 21.66 & 22.05 & 17.35 & 13.45\\
\textbf{LSTM+OL} & E & 11.41 & 21.36 & 22.47 & 17.60 & 13.48\\ \midrule
\textbf{NBEATS} & EM & 11.66 & 76.23 & 92.94 & 73.37 & 53.64\\
\textbf{NBEATS} & E & 11.71 & 87.35 & 106.98 & 83.10 & 61.62\\
\textbf{NBEATS} & ET & 12.30 & 84.76 & 110.95 & 77.96 & 59.28\\
\textbf{LSTM+OL} & ET & 12.49 & 38.26 & 33.48 & 23.08 & 14.34\\
\textbf{VAR} & E & 12.70 & 89.64 & 107.16 & 87.43 & 61.36\\
\textbf{VAR+OL} & E & 12.70 & 89.64 & 107.16 & 87.43 & 61.36\\
\textbf{VAR} & ET & 14.73 & 101.85 & 119.63 & 101.82 & 70.37\\
\textbf{VAR+OL} & ET & 14.73 & 101.85 & 119.63 & 101.82 & 70.37\\
\textbf{NBEATS} & ETM & 16.45 & 134.57 & 173.04 & 126.76 & 95.06\\
\textbf{LSTM} & ET & 17.82 & 181.81 & 224.16 & 172.39 & 123.94\\
\textbf{LSTM} & E & 18.01 & 180.27 & 220.82 & 171.09 & 125.25\\
\textbf{LSTM} & EM & 18.05 & 185.04 & 227.54 & 175.75 & 127.55\\
\textbf{LSTM} & ETM & 18.52 & 185.39 & 228.29 & 176.47 & 128.00\\
\textbf{XGBoost} & E & 19.12 & 188.56 & 231.48 & 179.20 & 131.04\\
\textbf{XGBoost} & ET & 19.16 & 189.34 & 232.74 & 180.14 & 130.90\\
\textbf{RF} & EM & 19.49 & 191.75 & 234.62 & 182.22 & 134.48\\
\textbf{XGBoost} & ETM & 21.36 & 192.14 & 235.06 & 182.23 & 134.75\\
\textbf{RF} & E & 21.39 & 192.45 & 235.33 & 182.87 & 134.85\\
\textbf{XGBoost} & EM & 21.81 & 191.20 & 234.22 & 180.25 & 134.80\\
\textbf{RF} & ET & 25.25 & 193.53 & 236.40 & 183.95 & 135.93\\
\textbf{RF} & ETM & 25.42 & 193.73 & 236.61 & 184.14 & 136.14\\
\textbf{CopyLastWeek} & E & 45.26 & 57.21 & 38.31 & 32.90 & 53.17\\
\textbf{CopyLastHour} & E & 53.66 & 50.62 & 60.68 & 40.08 & 65.93\\
\textbf{ES} & E & 53.66 & 50.62 & 60.68 & 40.08 & 65.93\\
\textbf{CopyLastDay} & E & 70.80 & 68.29 & 75.00 & 53.54 & 89.29\\
\textbf{VAR} & ETM & 198.03 & 145.25 & 154.38 & 166.35 & 105.84\\
\textbf{VAR+OL} & ETM & 198.03 & 145.25 & 154.38 & 166.35 & 105.84\\
\textbf{VAR} & EM & 200.42 & 146.02 & 151.40 & 167.79 & 101.00\\
\textbf{VAR+OL} & EM & 200.42 & 146.02 & 151.40 & 167.79 & 101.00\\

\bottomrule
\end{tabular}
\end{table}

\section{Results and Analysis}

Due to space limitations, only a selection of results will be presented here, given the extensive number of experiments conducted, totaling 
$ 3380 \textrm{ results} = $
$13 \textrm{ BCs} \; \times \: $
$5  \textrm{ periods} \; \times \: $
$4  \textrm{ feature sets} \; \times \: $
$(4 \textrm{ univariate}$
$\textrm{methods} \; + \: 9 \textrm{ multivariate methods} )$.
Instead, the complete results is publicly available here: \url{https://github.com/aprbw/OoD_Electricity_Forecasting_during_COVID-19}.
Specifically, the full results will be displayed for BC 8, characterized by its extended duration and wide standard deviation, enabling the identification of notable patterns and trends within the dataset.
Then, the analysis will be extended to other datasets to verify the generalizability of the findings and establish a more robust understanding of the implications of the research.

\subsection{Building Complex 8}

The results for all combinations of methods and feature sets for Building Complex 8 (BC 8) are displayed in Table \ref{tab:bc8}. For stochastic methods, the mean of ten replications, each with a different random seed, is given.

For BC 8, the continual learning approach (FSNet) outperforms other methods, and incorporating the mobility feature proves beneficial. This trend holds true across all five periods. A horizontal line in the table separates the methods that outperform the naive baselines in at least one period from those that do not. All methods above this line employ OL or CL, highlighting their importance during Out-of-Distribution periods.
Put differently, without the application of OL, these methods perform worse than the naive ones during Out-of-Distribution scenarios.
Note that linear models like VAR struggle to effectively leverage mobility data.

FSNet significantly outperforms the second-best method
(NBEATS + OL), showing an improvement ranging from 20\% to 40\%.
The addition of mobility data (transitioning from FSNet ET to FSNet EM) leads to a smaller but still noticeable improvement, ranging from 3\% to 10\%. 
We will quantify the improvements brought by mobility data, when such is the case, for other datasets in Section \ref{sec:quant_mob}.
Interestingly, FSNet ETM performs worse compared to the other FSNet variants, suggesting that the inclusion of additional features increase the risk of overfitting.

\begin{table}[htb]
\caption{
This table shows the best combinations of method and features for each BC.
MAE is the error metric, lower is better.
}\label{tab:best}
\begin{tabular}{@{}ccc|rrrrr@{}}
\toprule
\textbf{BC} & \textbf{Methods} & \textbf{Feat.} & \multicolumn{1}{c}{\textbf{PLD}} & \multicolumn{1}{c}{\textbf{LD1}} & \multicolumn{1}{c}{\textbf{IL1}} & \multicolumn{1}{c}{\textbf{LD2}} & \multicolumn{1}{c}{\textbf{IL2}} \\ \midrule
\textbf{1} & VAR & E & 8.04 & 8.76 & 11.18 & 9.83 & 7.92 \\
\textbf{2} & FSNet & ET & 2.37 & 1.20 & 1.04 & 0.80 & 1.57 \\
\textbf{3} & FSNet & EM & 15.63 & 10.33 & 9.59 & 7.15 & 12.32 \\
\textbf{4} & FSNet & EM & 16.04 & 9.48 & 11.06 & 6.99 & 11.80 \\
\textbf{5} & FSNet & ETM & 1.96 & 1.92 & 2.95 & 1.69 & 1.56 \\
\textbf{6} & FSNet & E & 14.71 & 11.19 & 11.32 & 9.50 & 9.85 \\
\textbf{7} & FSNet & E & 10.21 & 5.69 & 5.95 & 4.08 & 6.21 \\
\textbf{8} & FSNet & EM & 5.27 & 11.80 & 12.50 & 10.46 & 8.10 \\
\textbf{9} & FSNet & E & 4.68 & 2.31 & 3.56 & 1.93 & 4.31 \\
\textbf{10} & FSNet & E & 5.47 & 4.11 & 4.68 & 4.25 & 4.88 \\
\textbf{11} & FSNet & ET & 4.10 & 2.15 & 2.17 & 1.70 & 2.42 \\
\textbf{12} & FSNet & ETM & 0.22 & 0.26 & 0.47 & 0.27 & 0.20 \\
\textbf{13} & FSNet & E & 16.88 & 15.05 & 15.17 & 12.97 & 12.22 \\ \bottomrule
\end{tabular}
\end{table}

\subsection{Best over all BCs}

To assess the consistency of these trends across all BCs, we refer to Table \ref{tab:best}, which only displays the best combinations of methods and feature sets for each BC for brevity.
It is clear that FSNet is almost invariably the most effective method, thereby highlighting the crucial role of continual learning.
Although not explicitly shown in the table, it is important to note that the most effective combinations of features and methods remain practically consistent across all five periods.
This implies that FSNet remains the superior choice even during PLD.

While the benefits of adding extra features seem to vary, our findings indicate that mobility data provides an advantage in 5 out of 13 cases, and temperature data in 4 out of 13 instances. This suggests that, when feasible, it is always worthwhile to consider incorporating mobility data into the forecasting model as it may improve the forecast accuracy.

It is commonly assumed that out-of-distribution periods pose more significant challenges in forecasting. However, our analysis indicates a more nuanced picture. When comparing the forecasting accuracy across different time periods (horizontally on the table), we find that, for all BCs except 1, 6, 8, and 13, forecasting is most challenging during the Pre-Lockdown (PLD) period. This can be attributed to the decreased variability in electricity load during lockdown periods, as illustrated in Figure \ref{fig:vizabs}.

We can delve deeper into this aspect by assigning rankings to the time periods based on the difficulty faced by FSNet in forecasting for each BC, as presented in Table \ref{tab:best}. For instance, in the case of BC 1, the Inter-Lockdown 2 (IL2) period is ranked 1, indicating it is the easiest to forecast, as evidenced by the lowest error rate of 7.92. The Pre-Lockdown (PLD) period follows with a rank of 2, and so on, until the Inter-Lockdown 1 (IL1) period, which with a rank of 5, emerges as the most challenging for FSNet to forecast.

\begin{table}[htb]
\centering
\caption{Quantifying improvements of the best methods over the second best for every BC.}
\label{tab:quant_method}
\setlength\tabcolsep{5pt}
\begin{tabular}{@{}ccc|rrrrr@{}}
\toprule
\textbf{BC} & \textbf{Methods} & \textbf{Feat.} & \textbf{PLD} & \textbf{LD1} & \textbf{IL1} & \textbf{LD2} & \textbf{IL2} \\ \midrule \midrule

\multirow{3}{*}{1}  & VAR       		  & E  & 8.04  & 8.76  & 11.18 & 9.83   & 7.92  \\
				    & FSNet     		  & EM & 13.66 & 9.37  & 11.93 & 9.51   & 9.87  \\
				    & \textbf{$\Delta$\%} &    & \textbf{41\%} & \textbf{6\%}   & \textbf{6\%}   & -\textbf{3\%}   & \textbf{20\%}  \\ \midrule
\multirow{3}{*}{2}  & FSNet     		  & ET & 2.37  & 1.20  & 1.04  & 0.80   & 1.57  \\
				    & NBEATS+OL 		  & EM & 2.91  & 1.40  & 1.20  & 0.98   & 2.02  \\
				    & \textbf{$\Delta$\%} &    & \textbf{19\%} & \textbf{14\%}  & \textbf{13\%}  & \textbf{18\%}   & \textbf{22\%}  \\ \midrule
\multirow{3}{*}{3}  & FSNet     		  & EM & 15.63 & 10.33 & 9.59  & 7.15   & 12.32 \\
				    & VAR+OL    		  & E  & 23.64 & 14.96 & 12.04 & 10.33  & 20.14 \\
				    & \textbf{$\Delta$\%} &    & \textbf{34\%} & \textbf{31\%}  & \textbf{20\%}  & \textbf{31\%}   & \textbf{39\%}  \\ \midrule
\multirow{3}{*}{4}  & FSNet     		  & EM & 16.04 & 9.48  & 11.06 & 6.99   & 11.80 \\
				    & VAR+OL    		  & E  & 23.34 & 16.78 & 16.80 & 10.60  & 20.16 \\
				    & \textbf{$\Delta$\%} &    & \textbf{31\%} & \textbf{44\%}  & \textbf{34\%}  & \textbf{34\%}   & \textbf{41\%}  \\ \midrule
\multirow{3}{*}{5}  & FSNet     		  & ETM& 1.96  & 1.92  & 2.95  & 1.69   & 1.56  \\
				    & NBEATS+OL 		  & EM & 3.00  & 2.88  & 3.78  & 2.52   & 2.38  \\
				    & \textbf{$\Delta$\%} &    & \textbf{34\%} & \textbf{33\%}  & \textbf{22\%}  & \textbf{33\%}   & \textbf{34\%}  \\ \midrule
\multirow{3}{*}{6}  & FSNet     		  & E  & 14.71 & 11.19 & 11.32 & 9.50   & 9.85  \\
				    & VAR+OL    		  & E  & 24.15 & 12.65 & 13.65 & 11.48  & 16.35 \\
				    & \textbf{$\Delta$\%} &    & \textbf{39\%} & \textbf{12\%}  & \textbf{17\%}  & \textbf{17\%}   & \textbf{40\%}  \\ \midrule
\multirow{3}{*}{7}  & FSNet     		  & E  & 10.21 & 5.69  & 5.95  & 4.08   & 6.21  \\
				    & NBEATS+OL 		  & EM & 13.81 & 9.48  & 7.13  & 6.66   & 8.29  \\
				    & \textbf{$\Delta$\%} &    & \textbf{26\%} & \textbf{40\%}  & \textbf{17\%}  & \textbf{39\%}   & \textbf{25\%}  \\ \midrule
\multirow{3}{*}{8}  & FSNet     		  & EM & 5.27  & 11.80 & 12.50 & 10.46  & 8.10  \\
				    & NBEATS+OL 		  & EM & 7.02  & 16.41 & 17.91 & 17.41  & 13.80 \\
				    & \textbf{$\Delta$\%} &    & \textbf{25\%} & \textbf{28\%}  & \textbf{30\%}  & \textbf{40\%}   & \textbf{41\%}  \\ \midrule
\multirow{3}{*}{9}  & FSNet     		  & E  & 4.68  & 2.31  & 3.56  & 1.93   & 4.31  \\
				    & NBEATS    		  & EM & 5.47  & 25.73 & 11.18 & 22.39  & 8.39  \\
				    & \textbf{$\Delta$\%} &    & \textbf{14\%} & \textbf{91\%}  & \textbf{68\%}  & \textbf{91\%}   & \textbf{49\%}  \\ \midrule
\multirow{3}{*}{10} & FSNet     		  & E  & 5.47  & 4.11  & 4.68  & 4.25   & 4.88  \\
				    & VAR+OL    		  & E  & 8.43  & 5.42  & 5.80  & 6.37   & 7.49  \\
				    & \textbf{$\Delta$\%} &    & \textbf{35\%} & \textbf{24\%}  & \textbf{19\%}  & \textbf{33\%}   & \textbf{35\%}  \\ \midrule
\multirow{3}{*}{11} & FSNet     		  & ET & 4.10  & 2.15  & 2.17  & 1.70   & 2.42  \\
				    & XGBoost   		  & EM & 4.80  & 21.16 & 20.04 & 22.95  & 16.49 \\
				    & \textbf{$\Delta$\%} &    & \textbf{15\%} & \textbf{90\%}  & \textbf{89\%}  & \textbf{93\%}   & \textbf{85\%}  \\ \midrule
\multirow{3}{*}{12} & FSNet     		  & ETM& 0.22  & 0.26  & 0.47  & 0.27   & 0.20  \\
				    & NBEATS+OL 		  & ETM& 0.34  & 0.43  & 0.78  & 0.46   & 0.27  \\
				    & \textbf{$\Delta$\%} &    & \textbf{35\%} & \textbf{40\%}  & \textbf{40\%}  & \textbf{41\%}   & \textbf{26\%}  \\ \midrule
\multirow{3}{*}{13} & FSNet     		  & E  & 16.88 & 15.05 & 15.17 & 12.97  & 12.22 \\
				    & NBEATS+OL 		  & E+M& 22.09 & 20.04 & 18.36 & 17.16  & 12.74 \\
				    & \textbf{$\Delta$\%} &    & \textbf{24\%} & \textbf{25\%}  & \textbf{17\%}  & \textbf{24\%}   & \textbf{4\%}   \\

\bottomrule
\end{tabular}
\end{table}

\begin{table*}[htb]
\centering
\caption{
Comparison of MAE improvements brought by incorporating mobility features for five selected building complexes (BCs) where such enhancements are observed.
The table includes standard deviations ($\pm$) for comparison.
The row $\Delta$M represents the improvement achieved by adding mobility features, along with the corresponding percentage improvements.
Note that all results presented in the table are based on the FSNet model.
}\label{tab:deltaMob}
\begin{tabular}{@{}cc|rr|rr|rr|rr|rr@{}}
\toprule
BC & Feat. & \multicolumn{2}{c}{PLD} & \multicolumn{2}{c}{LD1} & \multicolumn{2}{c}{IL1} & \multicolumn{2}{c}{LD2} & \multicolumn{2}{c}{IL2} \\ \midrule  \midrule

\multirow{3}{*}{3} & E & 17.8796 & $\pm$0.4222 & 11.2482 & $\pm$0.3301 & 10.2727 & $\pm$0.3535 & 7.1119 & $\pm$0.1823 & 13.5849 & $\pm$0.3069 \\
 & EM & 15.6269 & $\pm$0.3064 & 10.3273 & $\pm$0.2541 & 9.5919 & $\pm$0.2462 & 7.1529 & $\pm$0.1300 & 12.3164 & $\pm$0.2794 \\
 & $\mathbf{\Delta}$\textbf{M} & \multicolumn{1}{c}{\textbf{2.2527}} & \textbf{12.60\%} & \textbf{0.9209} & \textbf{8.19\%} & \textbf{0.6807} & \textbf{6.63\%} & -\textbf{0.0409} & -\textbf{0.58\%} & \textbf{1.2686} & \textbf{9.34\%} \\ \midrule
\multirow{3}{*}{4} & E & 16.8746 & $\pm$0.2910 & 10.7414 & $\pm$0.2562 & 12.5059 & $\pm$0.2301 & 8.0558 & $\pm$0.1168 & 13.9173 & $\pm$0.2250 \\
 & EM & 16.0373 & $\pm$0.2219 & 9.4808 & $\pm$0.1886 & 11.0637 & $\pm$0.1255 & 6.9857 & $\pm$0.1160 & 11.8021 & $\pm$0.2188 \\
 & $\mathbf{\Delta}$\textbf{M} & \multicolumn{1}{c}{\textbf{0.8373}} & \textbf{4.96\%} & \textbf{1.2606} & \textbf{11.74\%} & \textbf{1.4422} & \textbf{11.53\%} & \textbf{1.0701} & \textbf{13.28\%} & \textbf{2.1152} & \textbf{15.20\%} \\ \midrule
\multirow{3}{*}{5} & ET & 2.4337 & $\pm$0.0145 & 2.4964 & $\pm$0.0238 & 3.4632 & $\pm$0.0457 & 1.9876 & $\pm$0.0170 & 1.8229 & $\pm$0.0187 \\
 & ETM & 1.9634 & $\pm$0.0241 & 1.9186 & $\pm$0.0380 & 2.9512 & $\pm$0.0500 & 1.6898 & $\pm$0.0183 & 1.5648 & $\pm$0.0306 \\
 & $\mathbf{\Delta}$\textbf{M} & \multicolumn{1}{c}{\textbf{0.4703}} & \textbf{19.32\%} & \textbf{0.5778} & \textbf{23.14\%} & \textbf{0.5121} & \textbf{14.79\%} & \textbf{0.2977} & \textbf{14.98\%} & \textbf{0.2581} & \textbf{14.16\%} \\ \midrule
\multirow{3}{*}{8} & E & 5.7669 & $\pm$0.0793 & 13.5915 & $\pm$0.3982 & 14.4428 & $\pm$0.3782 & 11.8113 & $\pm$0.2523 & 9.0967 & $\pm$0.1716 \\
 & EM & 5.2652 & $\pm$0.0548 & 11.7971 & $\pm$0.2779 & 12.5034 & $\pm$0.2087 & 10.4583 & $\pm$0.0536 & 8.1016 & $\pm$0.0706 \\
 & $\mathbf{\Delta}$\textbf{M} & \multicolumn{1}{c}{\textbf{0.5017}} & \textbf{8.70\%} & \textbf{1.7943} & \textbf{13.20\%} & \textbf{1.9395} & \textbf{13.43\%} & \textbf{1.3530} & \textbf{11.46\%} & \textbf{0.9952} & \textbf{10.94\%} \\ \midrule
\multirow{3}{*}{12} & ET & 0.2271 & $\pm$0.0015 & 0.2715 & $\pm$0.0053 & 0.5005 & $\pm$0.0087 & 0.2831 & $\pm$0.0043 & 0.2080 & $\pm$0.0022 \\
 & ETM & 0.2199 & $\pm$0.0018 & 0.2589 & $\pm$0.0028 & 0.4717 & $\pm$0.0024 & 0.2693 & $\pm$0.0019 & 0.2015 & $\pm$0.0008 \\
 & $\mathbf{\Delta}$\textbf{M} & \multicolumn{1}{c}{\textbf{0.0072}} & \textbf{3.18\%} & \textbf{0.0126} & \textbf{4.64\%} & \textbf{0.0289} & \textbf{5.77\%} & \textbf{0.0138} & \textbf{4.88\%} & \textbf{0.0065} & \textbf{3.11\%} \\ \midrule
 
\end{tabular}
\end{table*}

Averaging the rankings for each time period across all BCs, we find that LD2 period is the easiest to forecast, with an average ranking of 1.85. This is followed by LD1 period, with an average rank of 2.62. On the other hand, PLD period is identified as the most difficult to forecast, with the highest average rank of 4.15. The Inter-Lockdown periods (ILD1 and ILD2) secure average rankings of 2.62 and 2.77, respectively.
This analysis reveals that the periods of lockdown are the easiest to forecast, followed by the reopening periods, with the pre-lockdown periods proving to be the most challenging. However, it's crucial to note that these findings are primarily attributable to the use of FSNet. Recalling the findings presented in Table \ref{tab:bc8}, it's clear that methods without Online Learning (OL) fail to generalize during out-of-distribution periods.

\subsection{Quantifying Methods Improvements}

The magnitude of improvements achieved by FSNet across various BCs and time periods are substantial as shown in Table \ref{tab:quant_method}, making the display of standard deviations for comparison unnecessary. 
BC 11, in particular, showcases a remarkable impact, approaching nearly 100\% enhancement.
On average, across all BCs (excluding BC 1, which does not utilize FSNet) and time periods, FSNet demonstrates a substantial effect magnitude, with enhancements reaching 35\%.
These findings underscore the superiority of CL over ordinary online learning OL methods in achieving notable performance improvements in electricity load forecasting

In the table, NBEATS+OL demonstrated superior performance as the second-best method, whereas LSTM+OL did not yield favorable results. This highlights the importance of utilizing specialized deep learning models, such as NBEATS, that are specifically designed for time series analysis. It suggests that these models are more effective in capturing the complex temporal patterns and dynamics present in the electricity load data compared to more general models like LSTM.

Furthermore, our analysis demonstrates the significance of OL in the forecasting task. The majority of the second best methods incorporate OL. Interestingly, VAR+OL methods outperform LSTM+OL in serveral BCs, suggesting that simpler linear methods like VAR can achieve competitive performance, particularly when combined with online learning. This finding underscores the value of VAR+OL as a robust and interpretable baseline for similar forecasting tasks.

As mentioned before, this breakdown also shows that performance remains virtually consistent throughout the time periods, with the only exception being in BC 1 over LD2.

\subsection{Quantifying Mobility Improvements}
\label{sec:quant_mob}

Table \ref{tab:deltaMob} quantifies the improvements achieved by incorporating mobility features.
The enhancements range from a modest 3.11\% (in BC 12 over IL2) to a remarkable 23.14\% (in BC 5 over LD1).
These improvements are significant, with even the smallest improvement (in BC 12 over IL2) of 0.0065 being nearly three times the standard deviation of 0.0022.
This further emphasizes the importance of considering mobility data, as when it proves beneficial, the improvements can be substantial.
It is therefore recommended to explore the inclusion of mobility features if feasible.
As mentioned before, this breakdown also shows that performance remains virtually consistent throughout the time periods, with the only exception being in BC3 over LD2,

\section{Conclusions}

Our study offers several significant insights into the forecasting of energy demand, particularly in scenarios involving substantial and abrupt changes such as a global pandemic. These insights have far-reaching implications not only for future work in electricity load forecasting.

One of the crucial takeaways from our work is the importance of continual learning, particularly in handling out-of-distribution periods.
The continual learning strategy, as exemplified by FSNet, proves to be effective during times of drastic and unexpected changes.
As we witnessed during the COVID-19 pandemic, abrupt alterations in normal routines can significantly affect energy use patterns, which necessitates the use of models that can adapt quickly to new information.
This is a critical lesson, as the future is likely to bring more unpredictable situations that demand the robustness and adaptability provided by continual learning.

Another important lesson lies in the consideration of additional features, such as mobility and temperature data. The addition of these features leads to improvements in forecast accuracy for certain business complexes, suggesting that it is advantageous to incorporate such information when it is readily available. However, our study also highlights the risk of overfitting that comes with the inclusion of more features, reminding us of the need for careful management of model complexity.

Interestingly, our research challenges some common assumptions about the difficulty of forecasting. Contrary to expectations, the periods during lockdown were easier to forecast than the pre-lockdown periods. This finding underscores the importance of context and indicates that assumptions about prediction difficulty should always be tested empirically.

In conclusion, these insights will prove invaluable in guiding methodological decisions, shaping our understanding of prediction difficulty, and setting realistic expectations for model performance in forecasting electricity load and beyond.

\begin{acks}
We highly appreciate Centre for New Energy Technologies \href{https://c4net.com.au/}{(C4NET)} and Commonwealth Scientific and Industrial Research Organisation \href{www.csiro.au}{(CSIRO)} for their funding support and contributions during the project.
We would also like to acknowledge the support of Cisco's National Industry Innovation Network (NIIN) Research Chair Program.
This research was undertaken with the assistance of resources and services from the \href{https://nci.org.au/}{National Computational Infrastructure (NCI)}, which is supported by the Australian Government.
This endeavor would not have been possible without the contribution of Dr. Xinlin Wang and Dr. Mashud Rana.
\end{acks}

\bibliographystyle{ACM-Reference-Format}
\bibliography{1bib}


\end{document}